\documentclass[11pt,a4paper]{article}
\usepackage[hyperref]{acl2021}
\usepackage{times}
\usepackage{latexsym}

\usepackage{amsmath, amsfonts, amsthm, xspace, color}
\usepackage{booktabs} 

\usepackage{subfigure, multirow, tabularx} 
\usepackage{booktabs} 
\usepackage{enumitem}
\usepackage{flushend}
\usepackage[T1]{fontenc}
\usepackage{epstopdf}
\usepackage{balance}
\usepackage{graphicx}
\usepackage[noend,ruled,linesnumbered]{algorithm2e} 
\usepackage{microtype}
\usepackage{url}
\usepackage{wrapfig,lipsum}
\usepackage{makecell}
\usepackage{wrapfig}

\newcommand{\hide}[1]{} 

\newcommand{\etc}{\emph{etc.}\xspace} 
\newcommand{\ie}{i.e.\xspace} 
\newcommand{\eg}{\emph{e.g.}\xspace} 
\newcommand{\nop}[1]{}
\newcommand{\mquote}[1]{{``\emph{#1}''}}

\newtheorem{thm:def}{Definition}
\newtheorem{thm:eg}{Example}
\newtheorem{thm:lem}{Lemma}
\newtheorem{thm:obs}{Observation}
\newtheorem{thm:req}{Requirement}
\newtheorem{thm:prop}{Proposition}
\newtheorem{thm:principle}{Principle}
\newtheorem{thm:thm}{Theorem}
\newtheorem{thm:corollary}{Corollary}

\newcommand{\ours}{\mbox{\sc TEAMS}\xspace}

\definecolor{midnightgreen}{rgb}{0.0, 0.29, 0.33}
\definecolor{orange}{RGB}{255,127,0}

\usepackage{microtype}

\definecolor{OrangeRed}{rgb}{1.0, 0.27, 0.0}
\definecolor{midnightgreen}{rgb}{0.0, 0.29, 0.33}

\aclfinalcopy 
\def\aclpaperid{1919} 

\title{Training ELECTRA Augmented with Multi-word Selection}

\author{Jiaming Shen$^{\nabla\star}$, Jialu Liu$^{\Diamond}$, Tianqi Liu$^{\Diamond}$, Cong Yu$^{\Diamond}$, Jiawei Han$^{\nabla}$ \\
\small  $^{\nabla}$University of Illinois Urbana-Champaign, IL, USA, $^{\Diamond}$Google Research, NY, USA \\
\footnotesize $^{\nabla}$\{js2, hanj\}@illinois.edu $\quad$ $^{\Diamond}$\{jialu, tianqiliu, congyu\}@google.com $\quad$
}
\date{}

\begin{document}
\maketitle

{
\renewcommand{\thefootnote}{\fnsymbol{footnote}}
\footnotetext[1]{This work is done while interning at Google Research.  Corresponding Author: Jialu Liu.}
}

\begin{abstract}

Pre-trained text encoders such as BERT and its variants have recently achieved state-of-the-art performances on many NLP tasks.
While being effective, these pre-training methods typically demand massive computation resources.
To accelerate pre-training, ELECTRA trains a discriminator that predicts whether each input token is replaced by a generator.
However, this new task, as a binary classification, is less semantically informative. 
In this study, we present a new text encoder pre-training method that improves ELECTRA based on multi-task learning. 
Specifically, we train the discriminator to simultaneously detect replaced tokens and select original tokens from candidate sets.
We further develop two techniques to effectively combine all pre-training tasks: (1) using attention-based networks for task-specific heads, and (2) sharing bottom layers of the generator and the discriminator.
Extensive experiments on GLUE and SQuAD datasets demonstrate both the effectiveness and the efficiency of our proposed method. 

\end{abstract}

\section{Introduction}\label{sec:intro}

Contextualized representations from pre-trained text encoders have shown great power for improving many NLP tasks \cite{Rajpurkar2016SQuAD10,Wang2019GLUEAM,Wang2019SuperGLUEAS,Liu2019TextSW}.
Most pre-trained encoders, despite their variety, follow BERT~\cite{Devlin2018BERTPO} and adopt the masked language modeling (MLM) pre-training task which trains the model to recover the identities of a small subset of masked tokens.
Although being more effective than conventional left-to-right language model pre-training~\cite{Peters2018DeepCW, radford2018improving} due to capturing bidirectional information, MLM-based approaches~\cite{Liu2019RoBERTaAR,Joshi2019SpanBERTIP} can only learn from those masked tokens which are typically just 15\% of all tokens in the input sentences.

To address the low sample efficiency issue, ELECTRA~\cite{Clark2020ELECTRAPT} proposes a new pre-training task.
Specifically, it corrupts a sentence by replacing some tokens with plausible alternatives sampled from a generator and trains a discriminator to predict whether each token in the corrupted sentence is replaced or not. 
After pre-training ends, it throws away the generator and exports the discriminator for down-stream applications. 
As the discriminator can learn from all input tokens, ELECTRA is more sample efficient than previous MLM-based methods.
However, follow-up studies~~\cite{Xu2020MCBERTEL,ArocaOuellette2020OnLF} find this new replaced token detection task, as a binary classification, is often too simple to learn.
As a result, the discriminator output representations are insufficiently trained and encode inadequate semantic information. 

In this work, we propose a novel text encoder pre-training method \textbf{\textsc{TEAMS}} which stands for ``Training ELECTRA Augmented with Multi-word Selection''.
Compared with ELECTRA, our method also consists of a generator and a discriminator but they are equipped with different pre-training tasks.
For each masked position in the input sentence, the generator replaces the original token with an alternative token and samples a candidate set that consists of the original token and other $K$ non-original ones.
Then, we train the discriminator to simultaneously perform two tasks: (1) a \emph{multi-word selection} task in which the discriminator learns to select the original token from the sampled candidate set, and (2) a \emph{replaced token detection} task similarly defined in ELECTRA.
The first task, as a $(K+1)-$way classification on the masked positions, pushes the discriminator to differentiate ground truth tokens from other negative non-original ones. At the same time, the second task, with reduced task complexity, keeps the discriminator to achieve the same sample efficiency as ELECTRA. 

To further improve the performance and efficiency of our method, we introduce two techniques. 
The first one is using attention-based task-specific heads for discriminator multi-task pre-training. 
Different from previous studies~\cite{Liu2019MultiTaskDN, Sun2020ERNIE2A} that pass the last encoder layer outputs to different task heads, our method directly incorporates task-specific attention layers into the discriminator encoder. 
Such a design offers higher flexibility in capturing task-specific token dependencies in sequence and leads to significant performance boost. 
The second technique is to share the bottom layers of the generator and the discriminator. 
This technique reduces the number of parameters, saves computes, and serves as a form of regularization that stabilizes the training and helps the generalization. 

Combining above novelties all together, we train our models of various sizes and test their performance on the GLUE natural language understanding benchmark~\cite{Wang2019GLUEAM} and SQuAD question answering benchmark~\cite{Rajpurkar2016SQuAD10}.
We show that \ours substantially outperforms previous MLM-based methods and ELECTRA, given the same model size and pre-training data.
For example, our base-sized model, achieving 84.51 SQuAD 2.0 F1 score, outperforms BERT and ELECTRA by 8.34 and 2.99, respectively. 
Moreover, \ours-Base can outperform ELECTRA-Base++ using a fraction of computes.

\smallskip
\noindent \textbf{Contributions.}
The major contributions of this paper are summarized as follows:
(1) We propose a new text encoder pre-training method \ours that simultaneously learns a generator and a discriminator using multi-task learning.
(2) We develop two techniques, attention-based task-specific head and partial layer sharing, to further improve \ours performance.
(3) We conduct extensive experiments to verify the effectiveness of \ours on GLUE and SQuAD benchmarks\footnote{\small Code and pre-trained model weights are available at \url{https://bit.ly/acl21-teams-code}.}.
\section{Background}\label{sec:background}

In this section, we first discuss some related studies on pre-training text encoders.
Then, we introduce our notations and describe ELECTRA in details.

\subsection{Text Encoder Pre-training}\label{subsec:background_encoder_pretrain}

Current state-of-the-art natural language processing systems often rely on a text encoder to generate contextualized representations.
This text encoder is commonly pre-trained on massive unlabeled corpora using different self-supervised tasks.
\citet{Peters2018DeepCW} and \citet{radford2018improving} pre-train either a LSTM or a Transformer~\cite{Vaswani2017AttentionIA} using the standard language modeling task.
To further improve pre-trained models, more effective pre-training objectives have been developed, including masked language modeling and next sentence prediction in BERT~\cite{Devlin2018BERTPO}, permutation language modeling in XLNet~\cite{Yang2019XLNetGA}, masked span prediction in SpanBERT~\cite{Joshi2019SpanBERTIP}, sentence order prediction in StructBERT~\cite{Wang2020StructBERTIL}, and more.

Most pre-training methods demand massive amounts of computes, which limits their accessibilities and raises concerns about their environmental costs. 
To alleviate such issue, \citet{Gong2019EfficientTO} and \citet{Yang2020ProgressivelyS2} propose to accelerate BERT training by progressively stacking a shallow model to a deep model.
\citet{Gu2020OnTT} extend this idea by growing a low-cost model in different dimensions.
Along another line of work, \citet{Clark2020ELECTRAPT} propose a new pre-training task, named replaced token detection, that learns a text encoder to distinguish real input tokens from synthetically generated replacements.
Compared to BERT-style MLM pre-training in which only 15\% of tokens are utilized, ELECTRA can leverage \emph{all tokens} in input sentences and thus achieves better sample efficiency.
Following this idea, \citet{Xu2020MCBERTEL} propose a new pre-training task based on the multi-choice cloze test with a rejection option, and \citet{clark2020electric} connect ELECTRA with cloze modeling and pre-train the text encoder as an energy-based cloze model.
As our method is built upon ELECTRA, we discuss it in more detail below.
 
\subsection{ELECTRA}

ELECTRA jointly trains two models, a generator $G$ and a discriminator $D$.
Both models adopt the Transformer architecture as their backbones and map a sentence of $n$ tokens $\mathbf{x} = [x_1, \dots, x_n]$ to their corresponding contextualized representations $h(\mathbf{x}) = [h(\mathbf{x})_1, \dots, h(\mathbf{x})_n]$. 

The generator $G$ is trained using the masked language modeling (MLM) task.
Specifically, given an input sequence $\mathbf{x}$, it first randomly selects a few masked positions and replaces tokens at these positions with a special mask symbol \texttt{[MASK]}.
We denote this masked sequence as $\mathbf{x}^{M}$.
Then, the generator learns to predict the original identities of those masked-out tokens by minimizing the below MLM loss:
\begin{equation}\label{eq:mlm_loss}
\small
\mathcal{L}_{\text{MLM}}(\mathbf{x}; G) = \mathbb{E}\left(\sum_{i:x_{i}^{M}=\texttt{[MASK]}} -\log P_{G}(x_i|\mathbf{x}^{M}) \right),
\end{equation}
where $P_{G}(x_i|\mathbf{x}^{M})$ is the probability that $G$ predicts token $x_i$ appears in the masked position $i$ in $\mathbf{x}^{M}$, and the expectation is taken over the random draw of masked positions. 
More specifically, the generator calculates $P_{G}(x_i|\mathbf{x}^{M})$ by passing contextualized representations $h_{G}(\mathbf{x}^{M})$ into a softmax layer as follows:
\begin{equation}\label{eq:generator_softmax}
\small
P_{G}(x_i|\mathbf{x}^{M}) = \frac{\exp(e(x_i)^{T} h_{G}(\mathbf{x}^{M})_{i}) }{\sum_{x_i' \in V} \exp(e(x_i')^{T} h_{G}(\mathbf{x}^{M})_{i})  },
\end{equation}
where $e(x_i)$ is the embedding of token $x_i$ and $V$ denotes the vocabulary of all tokens.
Finally, for each masked position $i$, the generator samples one token $\hat{x}_{i} \sim P_{G}(\cdot|\mathbf{x}^{M})$ and replaces the original token $x_i$ with $\hat{x}_{i}$. 
We use $\mathbf{x}^{R}$ to denote this corrupted sentence with replaced tokens.

The discriminator $D$ learns to perform the replaced token detection (RTD) task that requires a model to predict whether each token in $\mathbf{x}^{R}$ is replaced or not.
In particular, ELECTRA adopts a sigmoid layer, on top of the discriminator output contextualized representations $h_{D}(\mathbf{x}^{R})$, to decide the probability that token $x^{R}_{i}$ matches the original token $x_{i}$ as follows:
\begin{equation}
\small
P_{D}(x^{R}_{i} = x_{i}) = \text{sigmoid}(w^{T}h_{D}(\mathbf{x}^{R})_i),
\end{equation}
where $w$ is a learnable parameter. 
The loss on $D$ is then defined as follows:
\small
\begin{align}\label{eq:rtd_loss}
         \mathcal{L}_{\text{RTD}}(\mathbf{x}, \mathbf{x}^{R}; & D) = \mathbb{E} \Big( \sum_{i: x^{R}_i = x_i}  -\log P_{D}(x^{R}_i = x_i)  \nonumber \\
        											       & + \sum_{i: x^{R}_i \neq x_i}  -\log (1-P_{D}(x^{R}_i = x_i) ) \Big).
\end{align}
\normalsize
Finally, the generator and discriminator are jointly learned based on losses in Eq.~(\ref{eq:mlm_loss}) and Eq.~(\ref{eq:rtd_loss}). 
After pre-training, ELECTRA throws out the generator and keeps only the discriminator for fine-tuning on downstream tasks.

 \begin{figure*}[!t]
   \centering
   \centerline{\includegraphics[width=0.98\textwidth]{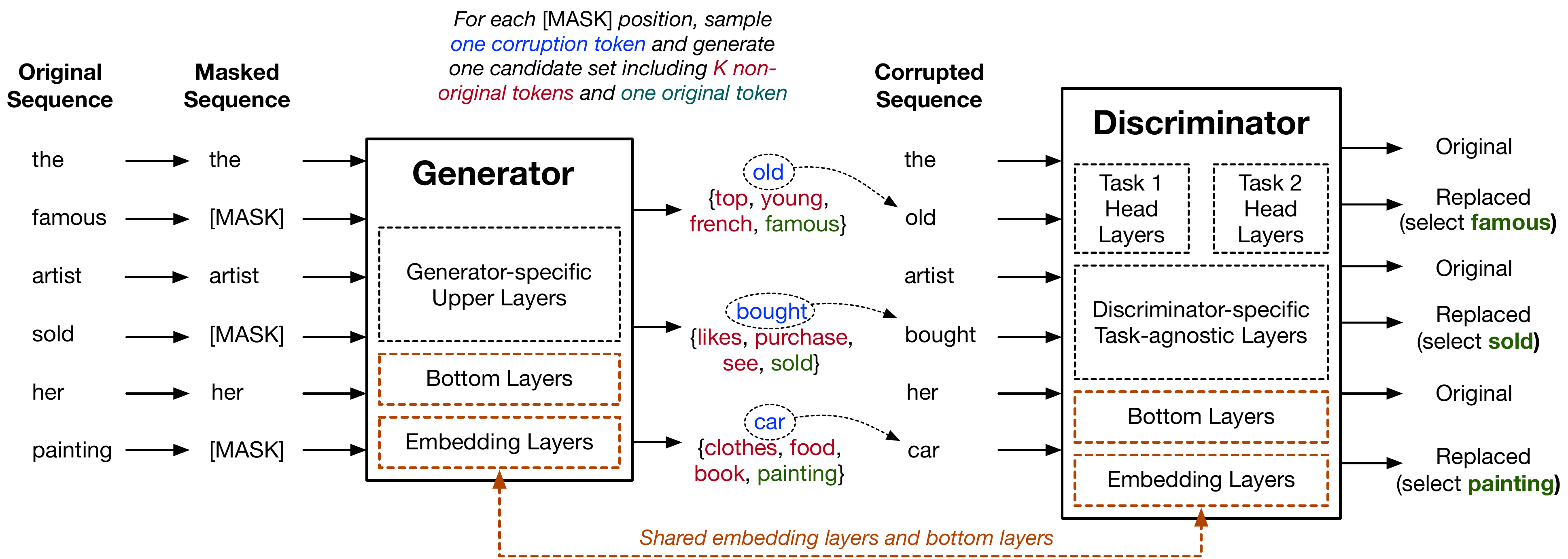}}
   \caption{The overview framework of \ours. For each masked position, the generator replaces its original token with a new one and outputs a candidate set consisting of the original token and another $K$ possible alternatives. The discriminator inputs the corrupted sentence and learns to (1) predict for every token whether it is replaced or not and (2) select the original token from the candidate set for each masked position.}
   \label{fig:framework}
 \end{figure*}
\section{The \textsc{TEAMS} Method}\label{sec:method}

In this section, we first introduce a new pre-training task named ``multi-word selection''.
Then, we present our \ours method with two techniques for performance improvements.

\subsection{Multi-word Selection Task}\label{subsec:sws}
To train a model on an input sequence $\mathbf{x}=[x_1, \dots, x_n]$ using the multi-world selection task, we first choose a random set of positions in this sequence, denoted as $\{i_1, \dots, i_{m}\}$ where $m$ is an integer between 1 and $n$.
Then, for each chosen position $i_j, j \in \{1, \dots, m\}$, we replace token $x_{i_j}$ with another token $\hat{x}_{i_j}$ and create a \emph{candidate set} $S_{i_j}$ that includes the original token $x_{i_j}$ and $K$ non-original ones. 
Following ELECTRA, we use $\mathbf{x}^{R}$ to denote the corrupted sentence with all tokens in chosen positions replaced. 
Finally, the model inputs the corrupted sentence and outputs a probability for selecting the original token $x_{i_j}$ from the candidate set $S_{i_j}$ as follows:
\begin{equation}
\small
P(x_{i_j}|\mathbf{x}^{R}, S_{i_j}) = \frac{\exp(e(x_{i_j})^{T} h(\mathbf{x}^{R})_{i_j}) }{\sum_{x^{'}_{i_j} \in S_{i_j}} \exp(e(x^{'}_{i_j})^{T} h(\mathbf{x}^{R})_{i_j})  },
\end{equation}
where $h(\mathbf{x}^{R})_{i_j}$ is the contextualized representation of token $\hat{x}_{i_j}$ from the model outputs.

Figure~\ref{fig:framework} shows a concrete example wherein a sequence of 6 tokens is given and its $2^{nd}$, $4^{th}$, and $6^{th}$ positions are chosen to be masked.
Take the $2^{nd}$ position as an example, the generator replaces the original token $x_{i_1}$=\mquote{famous} with another token $\hat{x}_{i_1}$=\mquote{old} and generates the candidate set $S_{i_1}$=\{\mquote{top}, \mquote{young}, \mquote{french}, \mquote{famous}\} which includes the original token $x_{i_1}$ and $K=3$ non-original alternatives.

We may view the multi-word selection task as a simplification of masked language modeling and a generalization of replaced token detection.
Asking the model to select the correct word from a candidate set rather than from the entire vocabulary, we can save more computes.
At the same time, being essentially a $(K+1)-$way classification problem, the multi-word selection task is more challenging than the replaced token detection task (which is a binary classification problem) and thus pushes the model to learn more semantic representations.
We describe how to generate the candidate set and present our entire method below.

\subsection{Multi-task Learning in \ours}
In \ours, we jointly train two transformer encoders, one as the generator network $G$ and the other as the discriminator network $D$.
Given a masked sequence $\mathbf{x}^{M}$, we use the generator $G$ to perform two tasks for each masked position $i_j$ in this sequence.
First, similar to ELECTRA, we sample one token $\hat{x}_{i_j} \sim P_{G}(\cdot|\mathbf{x}^{M})$ (c.f. Eq.~(\ref{eq:generator_softmax})) and obtain the corrupted sequence $\mathbf{x}^{R}$.
Second, we draw $K$ non-original tokens $\{x_{i_j}^{1}, \dots, x_{i_j}^{K}\}$ from $P_{G}(\cdot|\mathbf{x}^{M})$ without replacement\footnote{\small More discussions on other possible negative sampling strategies are presented in experiment section.} and construct the candidate set $S_{i_j} = \{x_{i_j}, x_{i_j}^{1}, \dots, x_{i_j}^{K}\}$.
Finally, we learn the generator $G$ using the standard masked language modeling task (c.f. Eq.~(\ref{eq:mlm_loss})). 

On the discriminator side, we train the discriminator network $D$ using two tasks --- replaced token detection (RTD) task and multi-word selection (MWS) task. 
Given a corrupted sentence $\mathbf{x}^{R}$ of length $n$, the discriminator will generate two sets of contextualized representations $\{h_{D}^{\text{RTD}}(\mathbf{x}^{R})_i\}    |_{i=1}^{n}$ and $\{h_{D}^{\text{MWS}}(\mathbf{x}^{R})_i\}   |_{i=1}^{n}$, one for each pre-training task.
For each position $i \in \{1, \dots, n \}$, we use $h_{D}^{\text{RTD}}(\mathbf{x}^{R})_i$ to calculate the probability that the token $\mathbf{x}_{i}^{R}$ is replaced as follows:
\begin{equation}\label{eq:our_rtd_prob}
\small
P_{D}(\mathbf{x}^{R}_{i} = \mathbf{x}_{i}) = \text{sigmoid}(w^{T}h_{D}^{\text{RTD}}(\mathbf{x}^{R})_i),
\end{equation}
and optimize the same RTD loss defined in Eq.~(\ref{eq:rtd_loss}).
For each masked position $i_j, j \in \{1, \dots, m\}$, we obtain the candidate set $S_{i_j}$ from generator outputs and use $h_{D}^{\text{MWS}}(\mathbf{x}^{R})_{i_j}$ to compute the probability of selecting the correct original token $x_{i_j}$ from this candidate set as follows:
\begin{equation}\label{eq:sws_prob}
\small
P_{D}(x_{i_j}|\mathbf{x}^{R}, S_{i_j}) = \frac{\exp(e(x_{i_j})^{T} h^{\text{MWS}}_{D}(\mathbf{x}^{R})_{i_j}) }{\sum_{x^{'}_{i_j} \in S_{i_j}} \exp(e(x^{'}_{i_j})^{T} h^{\text{MWS}}_{D}(\mathbf{x}^{R})_{i_j})  }.
\end{equation}
As the multi-word selection task is a multi-class classification problem, we define its loss function as follows:
\begin{equation}\label{eq:sws_loss}
\small
\mathcal{L}_{\text{MWS}}(\mathbf{x}, \mathbf{x}^{R}; D, \mathbb{S}) = \mathbb{E}\left(\sum_{j=1}^{m} -\log P_{D}(x_{i_j}|\mathbf{x}^{R}, S_{i_j}) \right),
\end{equation}
where $\mathbb{S} = \{S_{i_j} \}|_{j=1}^{m}$ is the collection of candidate sets at all masked positions.
Finally, we learn \ours by optimizing a combined loss as follows:
\begin{equation}\label{eq:final_loss}
{\small
\begin{aligned}
\min_{G, D} \Big( \mathcal{L}_{\text{MLM}}(\mathbf{x}; G) & + \lambda_1  \mathcal{L}_{\text{RTD}}(\mathbf{x}, \mathbf{x}^{R}; D) \\
	& + \lambda_2 \mathcal{L}_{\text{MWS}}(\mathbf{x}, \mathbf{x}^{R}; D, \mathbb{S}) \Big),
\end{aligned}
}
\end{equation}
where $\lambda_1$ and $\lambda_2$ are two loss balancing hyper-parameters. 
For the example sequence in Figure~\ref{fig:framework}, the discriminator needs to predict the tokens in $1^{st}$, $3^{rd}$, $5^{th}$ positions are not replaced, the tokens in $2^{nd}$, $4^{th}$, $6^{th}$ positions are replaced, and select tokens \mquote{famous}, \mquote{sold}, and \mquote{painting} in $2^{nd}$, $4^{th}$, $6^{th}$ positions, respectively.

After pre-training, we keep the discriminator network and fine-tune it for downstream applications\footnote{\small Empirically, we find that using the contextualized representations for the MWS task (\ie, $\{h_{D}^{\text{MWS}}(\mathbf{x}^{R})_i\}   |_{i=1}^{n}$) can achieve better fine-tuning performance.}.

\begin{figure}[!t]
   \centering
   \centerline{\includegraphics[width=0.48\textwidth]{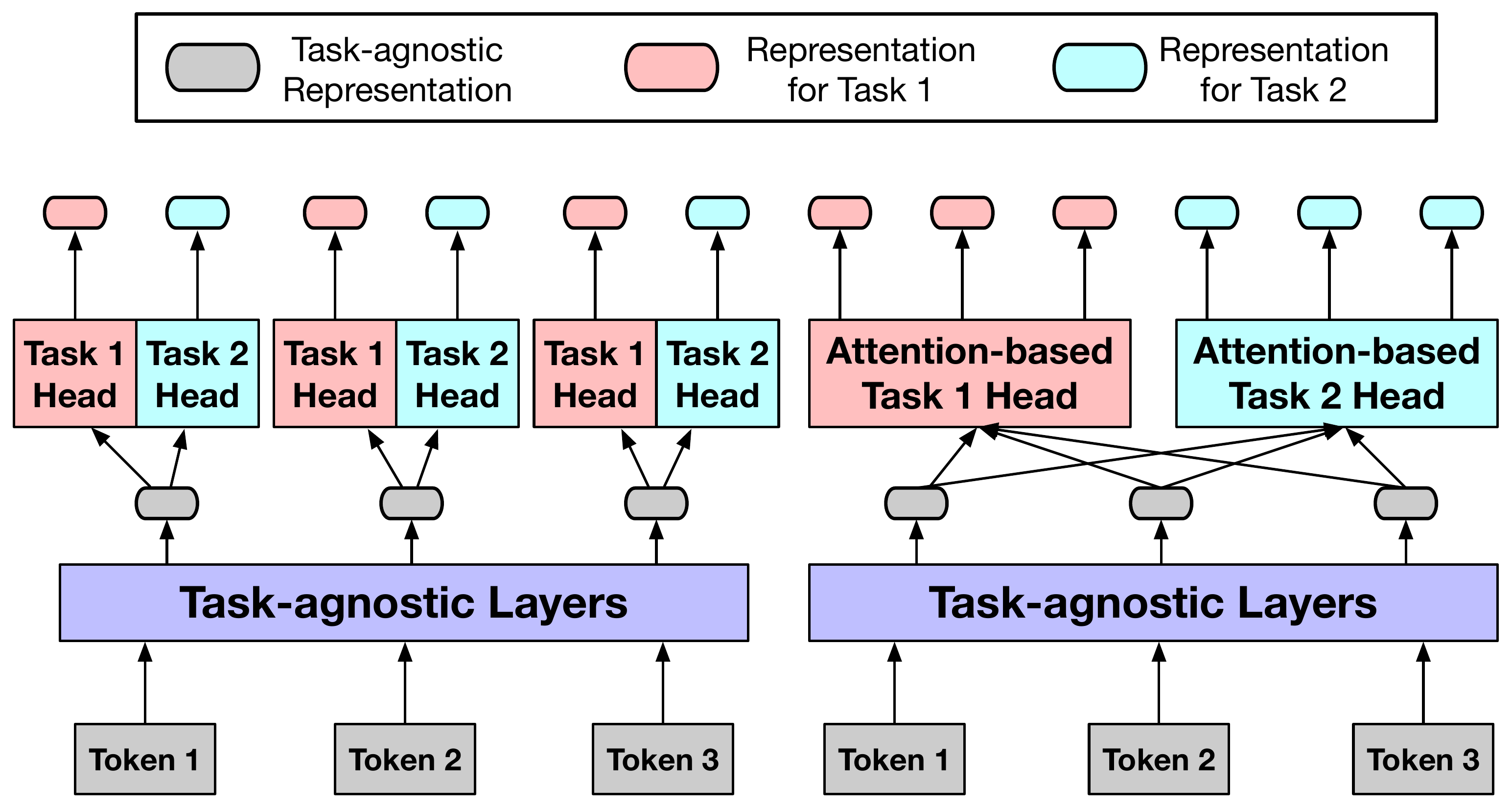}}
   \vspace{-0.1cm}
   \caption{Architectures for transforming task-agnostic representations to task-specific representations. (Left) Adding task-specific heads on each token separately. (Right) Using task-specific attention heads capture all token information holistically.}
   \label{fig:task_heads}
   \vspace{-0.1cm}
 \end{figure}

\smallskip
\noindent \textbf{Attention-based Task-specific Heads.}
One remaining question is how to generate two sets of \emph{task-specific} representations on the discriminator side.
Previous studies~\cite{Liu2019MultiTaskDN,Sun2020ERNIE2A,ArocaOuellette2020OnLF} achieve this goal by adding task-specific layers on top of each individual token, as shown in Figure~\ref{fig:task_heads} (Left). 
However, this approach does not model token dependencies within the task-specific layers.

In this work, we propose to use attention-based task-specific heads to capture global dependencies in sequences. 
Particularly, we design this attention head to be one transformer layer (\ie, a self-attention block followed by a fully connected network with one hidden layer). 
Since our discriminator also uses a transformer model to obtain each token's task-agnostic representation, we can merge one task head into the discriminator backbone.
From this perspective, we can generate different sets of \emph{task-specific} representations as follows.
First, we input the sequence to a transformer with $L$ layers and retrieve the final layer output representations for one task.
Then, we feed the output of an intermediate layer (\eg, the $(L-1)^{th}$ layer) into another transformer layer to obtain token representations for the second task.

\smallskip
\noindent \textbf{Partial Layer Sharing.}
ELECTRA has shown that tying the embedding layers of the generator and the discriminator can help improve the pre-training effectiveness.
Our study confirms this observation and finds that sharing some transformer layers of the generator and discriminator and can further boost the model performance.
More specifically, we design the generator to have the same ``width'' (\ie, hidden size, intermediate size and number of heads) as the discriminator and share the bottom half of all transformer layers between the generator and the discriminator.

\section{Experiments}\label{sec:exp}

\subsection{Experiment Setups}

 \begin{table*}[!t]
 	\centering
 	\scalebox{0.8}{
         \begin{tabular}{lcccccc}
         		\toprule
         	\multirow{2}{*}{\textbf{Method}} & \multirow{2}{*}{\textbf{Params}} & \multirow{2}{*}{\textbf{GLUE}} & \multicolumn{2}{c}{\textbf{SQuAD 1.1}} & \multicolumn{2}{c}{\textbf{SQuAD 2.0}} \\
                  & & & \bf EM & \bf F1 & \bf EM & \bf F1 \\
 		\midrule
		BERT-Small   		&14M	& 78.52 & 76.30	& 84.39	& 68.95	& 71.79 	\\
		ELECTRA-Small (Our reimplementation)	&14M	& 80.36	& 76.50	& 84.67	& 69.17	& 71.68	\\
		\ours-Small		&14M 	& \bf 80.70	& \bf 78.84	& \bf 86.40	& \bf 72.33	& \bf 75.24	\\
		\midrule
		BERT-Small++   				&14M		& 79.10 		& 76.48 & 84.75 & 68.37 & 71.01 \\	
		ELECTRA-Small++ (Our reimplementation)   		&14M		& 81.71		& 77.45 & 85.32 & 70.07 & 72.91 \\
		ELECTRA-Small++ (Public checkpoint re-evaluate)   &14M		& 81.24 		& 77.62  & 85.63	& 71.12	& 73.95	\\
		\ours-Small++				   		&14M	& \bf 81.99 		& \bf 78.94 & \bf 86.65 & \bf 72.11	& \bf 75.11	\\
		\midrule
		BERT-Base   		&110M	& 83.46 & 80.62 & 88.16 & 73.26 & 76.17 \\
		ELECTRA-Base (Our reimplementation)	&110M	& 84.63 &	83.87 & 90.64 & 78.59 & 81.52	\\
		\ours-Base 	&110M		& \bf 85.57 & \bf 85.21 & \bf 91.69 & \bf 81.59 & \bf 84.51 \\
		\midrule
		BERT-Base++   					&110M			& 84.26 	& 84.48 & 91.08 & 78.83 & 81.72 \\
		ELECTRA-Base++ (Our reimplementation)   			&110M			& 86.29	& 85.09 & 91.65 & 81.31 & 84.04	\\
		ELECTRA-Base++ (Public checkpoint re-evaluate)   &110M	& 87.13 	& 85.09 & 91.68 & 79.16	& 82.06 \\
		\ours-Base++   								&110M	& \bf 87.16	& \bf 86.05  & \bf 92.48	& \bf 82.73	& \bf 85.59	\\
		\midrule
		BERT-Large 									& 335M 	& 84.91 & 86.35  &92.61  & 82.19	& 84.78	 \\
		ELECTRA-Large (Our reimplementation)   			& 335M	& 89.20	& 88.79 & 94.50 & 86.02 & 88.72	\\
		ELECTRA-Large (Public checkpoint re-evaluate)   	& 335M	& 89.38 	& 88.76 & 94.49 & 86.79	& 89.56 \\
		\ours-Large & 335M 							 	& \bf 89.44 &\bf 88.86 & \bf 94.61 &\bf 87.08	& \bf 89.86	 \\
 		\bottomrule
          \end{tabular}
  	}
	\caption{Comparison results of \ours and baseline methods on GLUE and SQuAD datasets. All results are the medians of 15 fine-tuning runs with different initial random seeds. As ELECTRA original paper only releases the public checkpoints for Small++, Base++, and Large models, we can only report results for these three variants.}
 	\label{table:overall_results}
 	\vspace{-0.1cm}
 \end{table*}
 
\noindent \textbf{Pre-training Datasets.}
We use two datasets for model pre-training: 
(1) \textbf{WikiBooks}, which consists 3.3 Billion tokens from English Wikipedia and BooksCorpus~\cite{Zhu2015AligningBA}.
This is the same dataset used in BERT~\cite{Devlin2018BERTPO}.
(2) \textbf{WikiBooks++}, which extends WikiBooks dataset to 33 Billion tokens by including data from Gigaword~\cite{parker2011english}, ClueWeb~\cite{callan2009clueweb09}, and CommonCrawl~\cite{crawlcommon}.
The same dataset is used in XLNet~\cite{Yang2019XLNetGA} and ELECTRA~\cite{Clark2020ELECTRAPT}.

\smallskip
\noindent \textbf{Evaluation Datasets and Metrics.}
We evaluate all pre-trained models on the General Language Understanding Evaluation (GLUE) benchmark~\cite{Wang2019GLUEAM} and Stanford Question Answering (SQuAD) dataset~\cite{Rajpurkar2016SQuAD10}.
GLUE benchmark includes various tasks formatted as either single sentence classification (SST, CoLA) or sentence pair classification (\eg, RTE, MNLI, QNLI, MRPC, QQP, STS).
More details of each task are available in the Appendix Section A. 
SQuAD dataset requires models to select a text span from a given passage that answers a question. 
In SQuAD v1.1, the answers can always be located in the passage, while SQuAD v2.0 contains some questions unanswerable by the given passage. 

We compute Spearman correlation for STS, Matthews correlation for CoLA, accuracy for all other GLUE tasks, and report the GLUE score as the average of all 8 tasks.
For SQuAD, we use the standard evaluation metrics of Exact Match (EM) and F1 scores.
Since different random seeds may significantly affect fine-tuned model performances, we report the median of 15 fine-tuning runs from the same pre-trained model checkpoint for each result.
Unless stated otherwise, results are on the GLUE and SQuAD development sets.

\smallskip
\noindent \textbf{Model Hyper-parameters.}
We follow and evaluate \ours with different model sizes.
For \emph{small-sized} model, we set model hidden dimension to 256 and reduce token embedding dimension to 128.
The transformer in the discriminator network has 12 layers and each layer consists of 4 attention heads with the intermediate layer size 1024. 
For \emph{base-sized} model, we adopt the commonly used BERT-base configuration with 768 hidden dimension, 12 layers with 12 attention heads, and 3072 intermediate layer size. 
For \emph{large-sized} model, we use BERT-large configuration with 1024 hidden dimension, 24 layers with 16 attention heads, and 4096 intermediate layer size. 
Following~\cite{Clark2020ELECTRAPT}, we design the generator network size to be 1/2 of the discriminator network size for models of all sizes.  
For \ours, we set the loss balancing parameter $\lambda_1 = 5$, $\lambda_2 = 2$ (c.f., Eq.~(\ref{eq:final_loss})), and the number of sampled non-original tokens $K=5$ (c.f., Section~\ref{subsec:sws}).

During pre-training, we set the batch size to be 256 and the input sequence length to be 512 for both small-sized and base-sized models. 
We update small-sized models for 500K steps and base-sized models for 1M steps on the WikiBooks dataset. 
Moreover, we test the performance of each model when it is pre-trained for longer time with larger batch size using the WikiBooks++ dataset.
We use the suffix ``\emph{small++}'' to denote a small-sized model pre-trained for 2M steps with batch size 256, and the suffix ``\emph{base++}'' to denote a base-sized model pre-trained for 1M steps with batch size 1024. 
Finally, for large-sized models, we use batch size 2048 and pre-train the model for 1.76M steps.
All large-sized models and models with suffix ``\emph{++}'' are trained using the WikiBooks++ dataset.
More pre-training and fine-tuning details are included in the Appendix Section B and C.

\smallskip
\noindent \textbf{Model Implementations.}
For fair comparison, we implement all compared methods in TensorFlow 2 and evaluate their performances using the official pipeline in TensorFlow Model Garden\footnote{\small \url{https://github.com/tensorflow/models}.}.
In addition to our own implementations, we also report the performance of ELECTRA publicly released checkpoints\footnote{\small \url{https://github.com/google-research/electra}.}.
All models are trained on TPU v3.

\subsection{Experiment Results}

We validate the advantages of our proposed \ours method by comparing it with BERT~\cite{Devlin2018BERTPO} and ELECTRA~\cite{Clark2020ELECTRAPT}.
Table~\ref{table:overall_results} shows the comparison results on GLUE and SQuAD datasets.
We find that \ours can consistently outperform baseline models of the same size.
For example, compared to ELECTRA-Base, our \ours-Base improves SQuAD 2.0 performance from 78.59 to 81.59 and from 81.52 to 84.51 in terms of EM and F1 score, respectively.

\begin{figure}[!t]
   \centering
   \centerline{\includegraphics[width=0.49\textwidth]{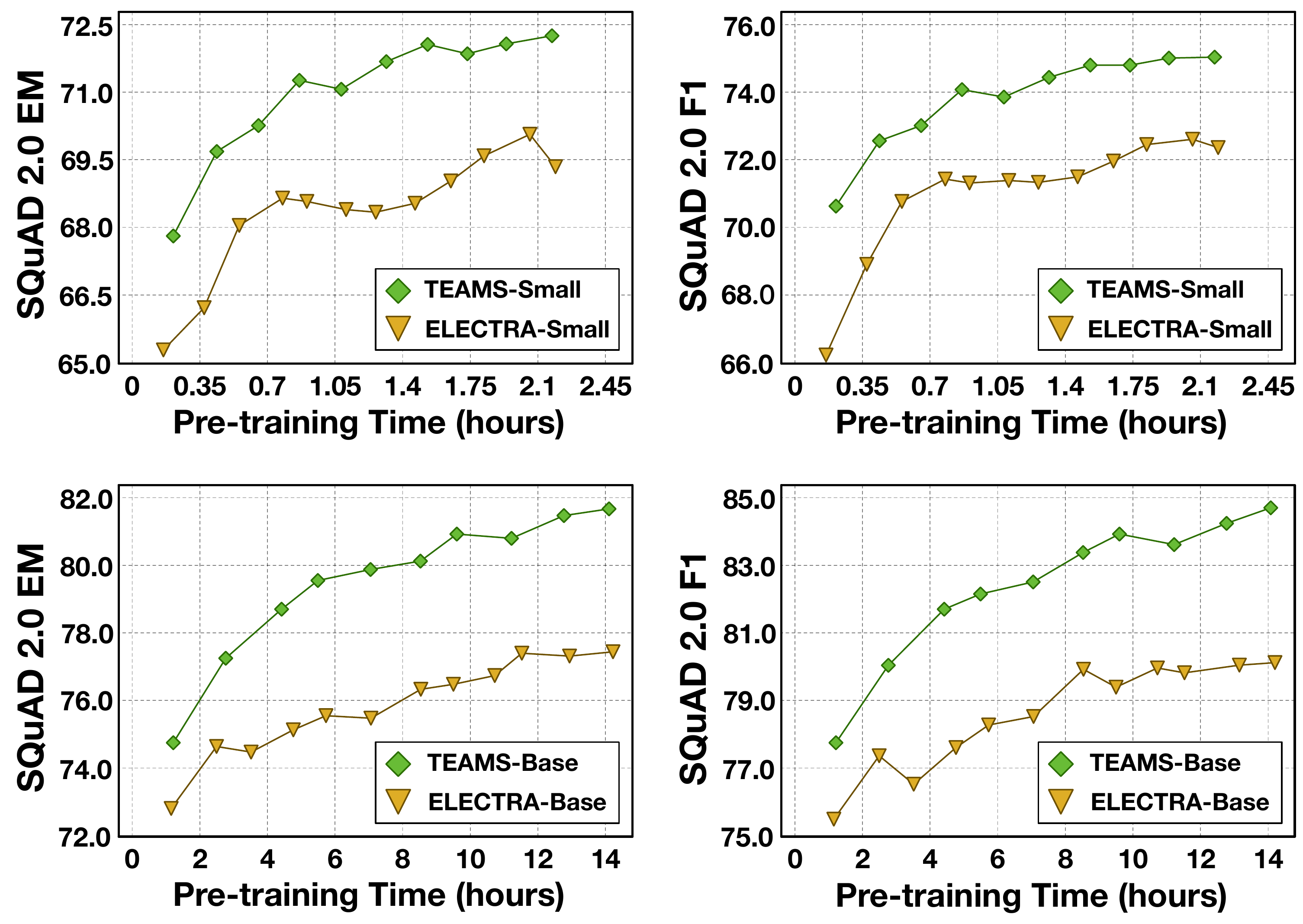}}
   \vspace{-0.1cm}
   \caption{Learning curves for small-sized and base-sized models.}
   \label{fig:learningCurve}
   \vspace{-0.1cm}
 \end{figure}

To further verify the performance improvements do not come from consuming more computations, we draw the learning curves of \ours-Small/Base and ELECTRA-Small/Base in Figure~\ref{fig:learningCurve}.
We observe that for both small-sized and base-sized models, our method can consistently outperform ELECTRA when trained for the some amount of time.
Moreover, on SQuAD datasets, \ours-Base can even outperform the ELECTRA-Base++ model that requires much more computation.

\subsection{Ablation Studies}\label{subsec:ablations}

We continue to evaluate the design of each component within \ours and test its sensitivity to some critical hyper-parameters.

\smallskip
\noindent \textbf{Effectiveness of Pre-training Tasks.}
We report the results of small-sized models learned using different pre-training tasks in Table~\ref{table:task_config}.
First, we can see that the model trained with multi-word selection (MWS) task can outperform the one learned using masked language modeling (MLM) task.
Second, on SQuAD datasets, we find that pre-training on only 15\% of masked tokens using MWS task is comparable with pre-training on all tokens using replaced token detection (RTD) task.
These observations demonstrate the effectiveness of our proposed MWS task.
Finally, we show that a text encoder pre-trained using both MWS and RTD tasks can outperform those learned using only single task.


 \begin{table}[!t]
 	\centering
 	\scalebox{0.665}{
         \begin{tabular}{lccccc}
         	\toprule
         	\multirow{2}{*}{\textbf{Pre-training Task(s)}} & \multirow{2}{*}{\textbf{GLUE}} & \multicolumn{2}{c}{\textbf{SQuAD 1.1}} & \multicolumn{2}{c}{\textbf{SQuAD 2.0}} \\
                  				& & \bf EM & \bf F1 & \bf EM & \bf F1 \\
 		\midrule
		Only MLM  (\ie, BERT) 			& 79.10 	& 76.48 	& 84.75 	& 68.37 	& 71.01 \\	
		Only RTD	 (\ie, ELECTRA)		& 81.71	& 77.45 	& 85.32 	& 70.07 	& 72.91 \\
		Only MWS 					& 79.65 	& 77.30	& 85.32	& 70.10	& 72.80    \\
		\midrule
		RTD + MWS (\ie, \ours) 		& \bf 81.99 		& \bf 78.94 & \bf 86.65 & \bf 72.11	& \bf 75.11	\\
 		\bottomrule
          \end{tabular}
  	}
	\caption{Effectiveness of multi-task pre-training for \emph{small++} models. ``MLM'', ``RTD'', and ``MWS'' stand for ``masked language modeling'', ``replaced token detection'', and ``multi-word selection'', respectively.}
 	\label{table:task_config}
 	\vspace{-0.1cm}
 \end{table}

\smallskip
\noindent \textbf{Task-specific Layer Designs.}
In \ours, we pre-train the discriminator network using multi-task learning and introduce the attention-based task-specific heads.
To verify the effectiveness of these attention-based task-specific heads, we train another model that uses the traditional feed forward network (FFN) as the task-specific head. 
Table~\ref{table:task_layer_analysis} shows the results. 
We can see that our model achieves better performances because the attention-based heads can effectively model the token dependencies in sequences.

We continue to study where to add these task-specific heads.
Currently, given a transformer with 12 layers, we treat its last layer output for one task and feed the $11^{th}$ layer output to a separate transformer layer to obtain representations for the second task\footnote{\small We can interpret the first 11 layers as \emph{task-agnostic} layers and view the $12^{th}$ layer and the newly introduced separate layers as two \emph{task-specific} heads.}.
An alternative design is to add two separate transformer layers (as two task-specific heads) directly on top of the last layer (\ie, the $12^{th}$ layer). 
As shown in Table~\ref{table:task_layer_analysis}, we find the latter design can slightly improve the model performance on SQuAD datasets but leads to a larger discriminator network with effectively 13 transformer layers and thus requires more computation during both pre-training and fine-tuning stages.

Finally, as our discriminator network will output two sets of contextualized representations, one for MWS task and the other for RTD task, we need to decide which set of representations to use in the fine-tuning stage.
Empirically, we find the representations for MWS task has better fine-tuning performance than the ones for RTD task, especially on the SQuAD datasets (c.f. Table~\ref{table:task_layer_analysis}).
This observation also confirms the effectiveness of our proposed MWS task as it produces representations capturing more fine-grained semantic information compared to the RTD task. 

 \begin{table}[!t]
 	\centering
 	\scalebox{0.64}{
         \begin{tabular}{lccccc}
         	\toprule
         	\multirow{2}{*}{\textbf{Method}} & \multirow{2}{*}{\textbf{GLUE}} & \multicolumn{2}{c}{\textbf{SQuAD 1.1}} & \multicolumn{2}{c}{\textbf{SQuAD 2.0}} \\
                  				& & \bf EM & \bf F1 & \bf EM & \bf F1 \\
 		\midrule
		ELECTRA-Small++  	& 81.71		& 77.45 & 85.32 & 70.07 & 72.91 \\
		\midrule
		\ours-Small++  								& 81.99 		&  78.94 &  86.65 &  72.11	&  75.11	\\
		\quad Use FFN task heads 				& 81.49 			& 78.18 & 86.35 & 72.00	& 74.90	\\   
		\quad Add task head on 12$^{th}$ layer 	& 81.29 			& 79.08 & 86.66 & 72.47	& 75.31	\\   		
		\quad Use RTD task head outputs			& 81.83 			& 77.72 & 85.80 & 69.56	& 72.57	\\   
 		\bottomrule
          \end{tabular}
  	}
	\caption{Analysis of task-specific layers and exported representations for \emph{small++} models. Please refer to Section~\ref{subsec:ablations} for detailed descriptions of each method.}
 	\label{table:task_layer_analysis}
 	\vspace{-0.1cm}
 \end{table}

\smallskip
\noindent \textbf{Partial Layer Sharing.}
Table~\ref{table:partial_layer} reports the results of our models with different levels of parameter sharing between the generator and the discriminator. 
First, we can see that tying all generator layers with discriminator layers results in significant performance drops, as such a binding restricts the model representation power. 
Second, we find that compared to no weight sharing, our design of partial layer tying can improve the model performance.
One possible explanation is that such layer tying serves as an implicit form of regularization and forces the shared transformer layers to capture useful information for both generator pre-training task (\ie, MLM) and discriminator pre-training tasks (\ie, RTD and MWS).

 \begin{table}[!t]
 	\centering
 	\scalebox{0.72}{
         \begin{tabular}{lccccc}
         	\toprule
         	\multirow{2}{*}{\textbf{Method}} & \multirow{2}{*}{\textbf{GLUE}} & \multicolumn{2}{c}{\textbf{SQuAD 1.1}} & \multicolumn{2}{c}{\textbf{SQuAD 2.0}} \\
                  				& & \bf EM & \bf F1 & \bf EM & \bf F1 \\
 		\midrule
		ELECTRA-Small++  			& 81.71		& 77.45 & 85.32 & 70.07 & 72.91 \\
		\ours-Small++				& \bf 81.99 		& \bf 78.94 & \bf 86.65 &  72.11	&  75.11	\\
		\quad Full Tie 				& 80.57 		& 77.75 & 85.76 & 69.93	& 72.82	\\   
		\quad No Tie 				& 81.65 		& 78.42 & 86.32 & \bf 72.73	& \bf 75.73	\\   
		\midrule
		ELECTRA-Base++  				& 86.29	& 85.09 & 91.65 & 81.31 & 84.04	\\
		\ours-Base++					& \bf 87.16	& \bf 86.05  & \bf 92.48	& \bf 82.73	& \bf 85.59	\\
		\quad No Tie 					& 86.63 	& 85.51 & 91.98 & 80.72	& 83.60	\\   
 		\bottomrule
          \end{tabular}
  	}
	\caption{Effect of sharing generator and discriminator bottom layers for \emph{small++} and \emph{base++} models. ``Full Tie'' and ``No Tie'' stand for tying all or none of generator layers with the discriminator, respectively.}
 	\label{table:partial_layer}
 	\vspace{-0.1cm}
 \end{table}

\begin{figure}[!t]
   \centering
   \centerline{\includegraphics[width=0.49\textwidth]{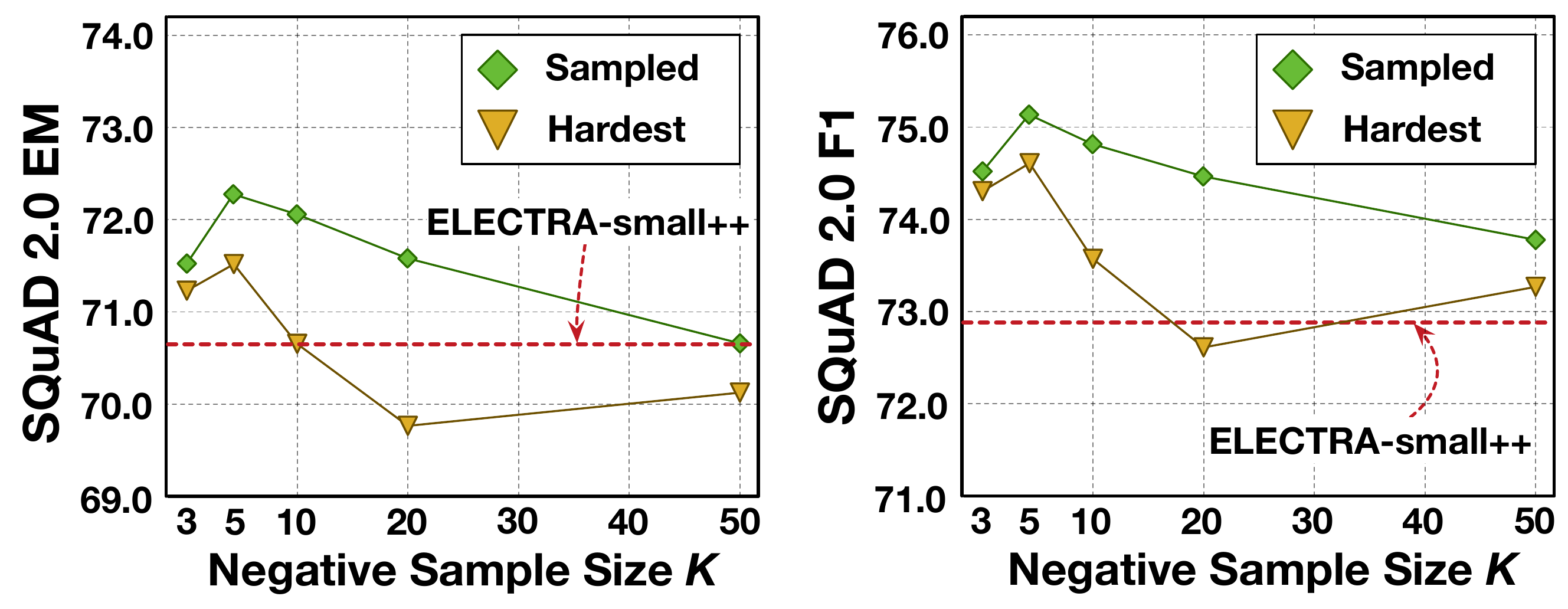}}
   \vspace{-0.1cm}
   \caption{Analysis of negative sampling strategies and negative sample sizes for \emph{small++} models. Based on the generator output distribution, we either perform weighted sampling without replacement $K$ times (``Sampled'') or select $K$ most likely non-original tokens (``Hardest'') for each masked position.}
   \label{fig:topK}
   \vspace{-0.1cm}
 \end{figure}

\smallskip
\noindent \textbf{Sampling Strategy and Negative Sample Size.}
To use the multi-world selection task for pre-training, we need to first obtain a set of negative samples (\ie, non-original tokens) for each masked position in a sequence.
In this study, we test two strategies to generate $K$ negative samples for each masked position.
Given the generator output probability distribution for a target position, we can either sample from this distribution $K$ times without replacement or directly select $K$ non-original tokens with the highest probabilities.
We denote these two approaches as ``Sampled'' and ``Hardest'', respectively, and report the results in Figure~\ref{fig:topK}.
First, we can see that performing repeated sampling is a better strategy than always selecting those hardest samples. 
One possible reason is that the ``Sampled'' strategy can generate more diverse negative samples and thus helps model to generalize\footnote{\small A similar result is also witnessed in~\cite{Shen2019MiningES} and thus we adopt the ``Sampled'' approach in this study.}.
Second, we notice that increasing $K$ over 5 will somewhat hurt the model performance. 
One reason is that a larger $K$ causes a higher probability of including false negative examples.
Finally, we find that for a wide range of $K$ from 3 to 50, our method can outperform ELECTRA, which further demonstrates the effectiveness of multi-word selection task.

\begin{figure}[!t]
   \centering
   \centerline{\includegraphics[width=0.48\textwidth]{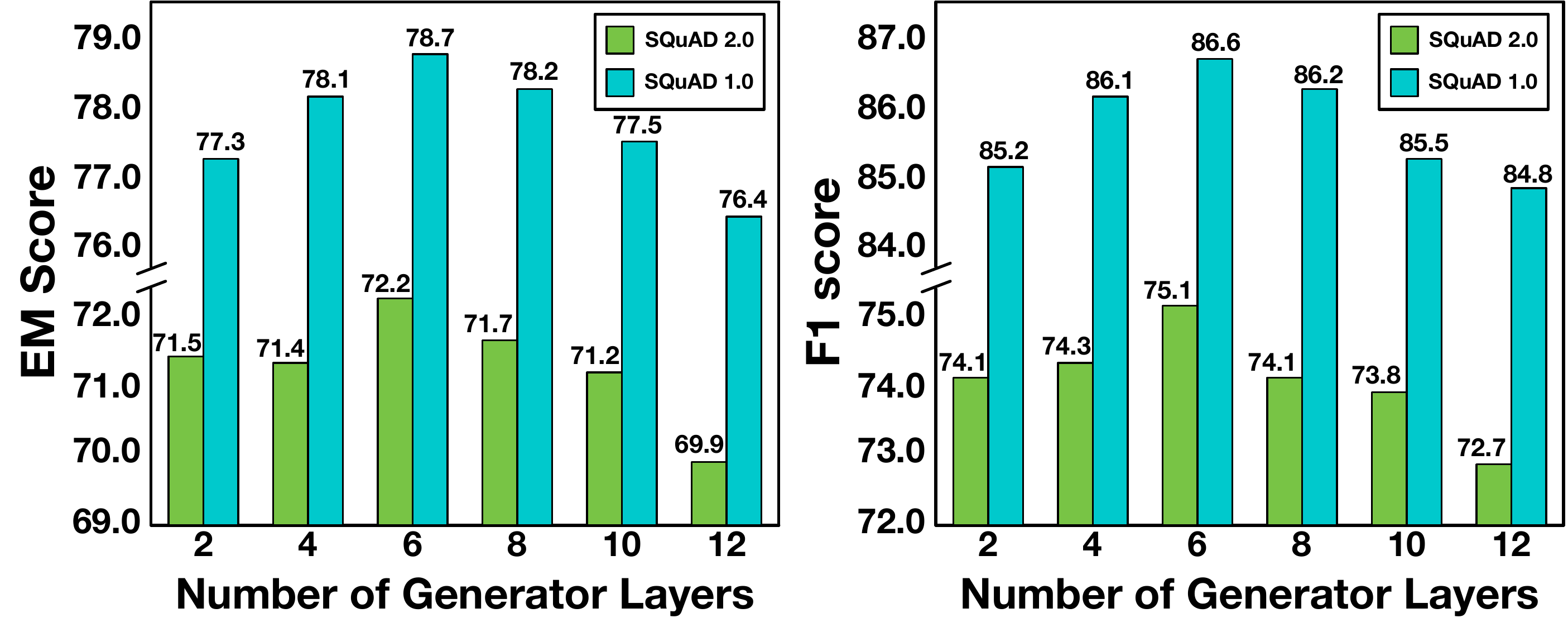}}
   \vspace{-0.1cm}
   \caption{Analysis of the number of generator layers in \ours-small++ models.}
   \label{fig:genLayers}
   \vspace{-0.1cm}
 \end{figure}

\smallskip
\noindent \textbf{Generator Size.}
We test how the size of generator affects the model performance by varying the number of transformer layers in the generator.
For all tested models, we tie the bottom half of generator with the discriminator. 
Figure~\ref{fig:genLayers} reports the results.
We find that the performance first increases as the generator size increases until it reaches about half of the discriminator size and then starts to decrease when we further increase the generator size. 
The same results also hold for base-sized models.

\section{Related Work}\label{sec:related_work}
Besides the general language pre-training work we discussed in Section~\ref{subsec:background_encoder_pretrain}, this study is particularly related to methods that apply multi-task learning~\cite{caruana1997multitask,Ruder2017AnOO,Shen2018MultiTaskLF} to language representation learning. 
An early study~\cite{Liu2019MultiTaskDN} proposes to simultaneously fine-tune a pre-trained BERT model to perform multiple natural language understanding tasks and achieves promising results on the GLUE dataset.
\citet{Sun2020ERNIE2A} continue this line of work and propose to push the multi-task learning to the model pre-training stage.
Specifically, they use a continual multi-task learning framework that incrementally builds and inserts seven auxiliary tasks (\eg, masked entity prediction, sentence distance prediction, \etc.) to the text encoder.
More recently, \citet{ArocaOuellette2020OnLF} extend this idea to incorporate fourteen auxiliary tasks and identify six tasks are particularly useful.
While achieving inspiring performance, these studies all assume the MLM pre-training task must present and just combine MLM with additional tasks.
In this paper, we relax this assumption and combine our new multi-word selection task with the replace token detection task for effective pre-training.


\section{Conclusions and Future Work}\label{sec:conclusion}

This work presents a new text encoder pre-training method that simultaneously learns a generator and a discriminator using multi-task learning. 
We propose a new pre-training task, multi-word selection, and combine it with previous pre-training tasks for efficient encoder pre-training. 
We also develop two techniques, attention-based task-specific heads and partial layer sharing, to further improve pre-training effectiveness.
Extensive experiments on GLUE and SQuAD datasets demonstrate our \ours method can consistently outperform previous state-of-the-arts methods.
In the future, we plan to explore how other auxiliary pre-training tasks can be integrated into our framework.
Moreover, we consider extending our pre-training method to text encoders with other architectures such as those based on dynamic convolution and sparse attention.
Finally, being orthogonal to this study, distillation techniques could be applied to further compress our pre-trained encoders into smaller models for faster inference speeds.

\section*{Acknowledgement}
We thank Hongkun Yu, You (Will) Wu from Google Research, Xiaotao Gu, Yu Meng from University of Illinois at Urbana-Champaign, and Richard Pang from New York University for providing valuable comments and discussions. Also, we would like to thank anonymous reviewers for valuable feedback.

\section*{Broader Impact Statement}

Recent years have witnessed the great success of pre-trained text encoders in lots of NLP applications such as text classification, question answering, text retrieval, dialogue system, etc.
This paper presents a new pre-training method \ours that learns a text encoder with better performance using lower training cost.
Therefore, on the positive side, our work has the potentials to benefit all downstream applications that leverage a pre-trained text encoder, especially those applications with limited computation resources.
On the negative side, \ours, as one specific pre-training method, could still face the generic issues for all language pre-training work.
For example, the pre-training large corpora, collected from the internet, may include abusive language usages and fail to capture the cultures that have smaller linguistic footprints online.

\bibliographystyle{acl_natbib}
\bibliography{cited.bib}


\newpage
\appendix


\section{GLUE Details}\label{app:glue}

The original GLUE benchmark~\cite{Wang2019GLUEAM} contains 9 natural language understanding datasets.
We describe them below:
\begin{itemize}[leftmargin=*]
    \item \textbf{MNLI:} The Multi-genre Natural Language Inference Corpus~\cite{Williams2018ABC} contains 393K training sentence pairs with textual entailment annotations. Given a premise sentence and a hypothesis sentence, a model needs to predict whether the premise entails the hypothesis, contradicts the hypothesis, or neither. 
    \item \textbf{RTE:} Recognizing Textual Entailment~\cite{Giampiccolo2007TheTP} dataset is similar to MNLI and contains 2.5K sentence pairs with binary entailment annotations (entailment or contradiction).     
    \item \textbf{QNLI:} Question Natural Language Inference dataset is a binary sentence pair classification dataset constructed from SQuAD~\cite{Rajpurkar2016SQuAD10}. It contains 108K training sentence pairs and requires a model to predict whether a context sentence contains the answer to a question sentence. 
    \item \textbf{CoLA:} Corpus of Linguistic Acceptability~\cite{Warstadt2018NeuralNA}. This dataset includes 8.5K training sentences annotated with whether it is a grammatical English sentence.
    \item \textbf{SST:} Stanford Sentiment Treebank~\cite{Socher2013RecursiveDM} dataset contains 67K sentences from movie reviews and their corresponding binary sentiment annotations.
    \item \textbf{MRPC:} Microsoft Research Paraphrase Corpus~\cite{Dolan2005AutomaticallyCA} includes 3.7K sentence pairs from online news sources. The task is to predict whether two sentences are semantically equivalent or not.
    \item \textbf{STS:} Semantic Textual Similarity~\cite{Cer2017SemEval2017T1} benchmark contains 5.8K training sentence pairs. The task is to predict the similarity score of two sentences from 1 to 5.
    \item \textbf{QQP:} Quora Question Pairs~\cite{QQP} dataset includes 364K question pairs sampled from the community question-answering website Quora. Models are trained to predict whether a pair of questions are semantically equivalent.    
    \item \textbf{WNLI:} Winograd NLI~\cite{Levesque2011TheWS} is a small natural language inference dataset. However, as GLUE official website\footnote{\small \url{https://gluebenchmark.com/faq}} indicates there are some issues during its construction process, we follow previous studies~\cite{Clark2020ELECTRAPT,Jiang2020ConvBERTIB} and exclude this dataset for fair comparisons. 
\end{itemize}

\section{Pre-training Details}\label{app:pre_train}

For the pre-training architecture configurations, we mostly use the same hyper-parameters as BERT and ELECTRA.
To generate masked positions, we follow BERT and duplicate training data 40 times so each sequence is masked in 40 different ways. 
We find this static masking strategy performs similar to the dynamic masking strategy in ELECTRA, while being easier to implement and has less computation overhead.
Besides, we notice a mask percentage of 15 works well for all models and thus do not increase it to 25 for large-size models as suggested in ELECTRA.
We set $\lambda_1$ and $\lambda_2$, the loss balancing parameters to 5 and 2, respectively, to ensure different loss terms are of the same scale. 
For small-size and base-size models, we search for the learning rate out of \{1e-4, 2e-4, 3e-4, 5e-4\}, batch size from \{128, 256, 512, 1024\}, and training steps from \{500K, 1M, 1.5M, 2M\}.
For large-size models, we search for the learning rate out of \{1e-4, 2e-4, 3e-4, 5e-4\} and batch size from \{1024, 2048\}.
Also, we select the generator size out of \{1/4, 1/3, 1/2\} in early experiments. 
The best configurations are reported in the main text and we perform no other hyper-parameter tuning.
The full set of hyper-parameters are listed in Table~\ref{table:supp_pre_train}.

 \begin{table*}[!t]
 	\centering
 	\scalebox{0.9}{
         \begin{tabular}{lccccc}
         	\toprule
		Hyper-parameter & Small/Small++ & Base/Base++ & Large \\
		\midrule
		Number of Layers & 12 & 12 & 24 \\
		Embedding Dim. & 128 & 768 & 1024 \\
		Hidden Dim. & 256 & 768 & 1024 \\
		Intermediate Layer Dim. & 1024 & 3072 & 4096 \\
		Number of Attention Heads & 4 & 12 & 16 \\
		Attention Head Dim. & 64 & 64 & 64 \\
		Generator Size (Multiplier for Number of Layers) & 1/2 & 1/2 & 1/2 \\
		Mask Percentage & 15 & 15 & 15 \\
		Learning Rate Decay & Linear & Linear & Linear \\
		Warmup Steps & 10000 & 10000 & 10000 \\
		Learning Rate & 5e-4 & 2e-4/3e-4 & 2e-4 \\
		Adam $\epsilon$ & 1e-6 & 1e-6 & 1e-6 \\
		Adam $\beta_1$ & 0.9 & 0.9 & 0.9 \\
		Adam $\beta_2$ & 0.999 & 0.999 & 0.999 \\
		Attention Dropout & 0.1 & 0.1 & 0.1 \\
		Dropout & 0.1 & 0.1 & 0.1 \\
		Weight Decay & 0.01 & 0.01 & 0.01 \\
		Batch Size  & 256/256 & 256/1024 & 2048 \\
		Train Steps & 500K/2M & 1M/1M & 1.76M \\
 		\bottomrule
          \end{tabular}
  	}
	\caption{Pre-training hyper-parameters.}
 	\label{table:supp_pre_train}
 	\vspace{-0.1cm}
 \end{table*}

\section{Fine-tuning Details}\label{app:fine_tune}
For fair comparisons, we fine-tune all pre-trained checkpoints using the official pipeline in TensorFlow Model Garden\footnote{\small \url{https://github.com/tensorflow/models}} and report the median of 15 fine-tuning runs. 
We do not include layer-wise learning-rate decay.
We search for the learning rate from \{1e-5, 3e-5, 5e-5, 8e-5, 1e-4\}, batch size from \{32, 48\}, and training epoch from \{2, 3, 5\}. For GLUE tasks, best evaluation scores during fine-tuning are reported. For SQuAD, scores at the end of fine-tuning are reported.
The full set of hyper-parameters are listed in Table~\ref{table:supp_fine_tune}.

 \begin{table*}[!t]
 	\centering
 	\scalebox{0.9}{
         \begin{tabular}{ll}
         	\toprule
		Hyper-parameter & Value \\
		\midrule
		Learning Rate &  1e-4 in Small/Small++, 3e-5 in Base/Base++/Large for GLUE \\ 
		              &  8e-5 in Small/Small++, 5e-5 in Base/Base++, and 3e-5 in Large for SQuAD 1.1/2.0 \\
		Learning Rate decay & Linear \\
		Warmup fraction & 0.1 \\
		Adam $\epsilon$ & 1e-6 \\
		Adam $\beta_1$ & 0.9 \\
		Adam $\beta_2$ & 0.999 \\
		Dropout & 0.1 \\
		Batch Size  & 32 for GLUE, 48 in Small/Small++/Large and 32 in Base/Base++ for SQuAD 1.1/2.0 \\
		Training Epochs  & 5 for GLUE, 5 in Small/Small+ and 2 in Base/Base++/Large for SQuAD 1.1/2.0 \\
 		\bottomrule
          \end{tabular}
  	}
	\caption{Fine-tuning hyper-parameters.}
 	\label{table:supp_fine_tune}
 	\vspace{-0.1cm}
 \end{table*}


\end{document}